\documentclass{article} 

\newcommand{\accuracyTable}{{
\begin{table*}[t]
\centering
\captionof{table}{
PPL(\textcolor{mygreen}{$\downarrow$}) and Zero-shot(\textcolor{mygreen}{$\uparrow$}) performance of LLaMA-7B with Basis Sharing and baselines under 20\% to 50\% compression ratio on three language modeling datasets and seven common sense reasoning datasets. The $\mS$ of all tasks is obtained with the dataset WikiText-2.
}
\vspace{-2mm}
\resizebox{1\textwidth}{!}{
\begin{tabular}{ccccccccccccc}
\midrule
\multicolumn{1}{c|}{\textsc{Ratio}}       & \multicolumn{1}{c|}{\textsc{Method}}    & WikiText-2\textcolor{mygreen}{$\downarrow$} & PTB\textcolor{mygreen}{$\downarrow$} & \multicolumn{1}{c|}{C4\textcolor{mygreen}{$\downarrow$}} & Openb. & ARC\_e & WinoG. & HellaS. & ARC\_c & PIQA & MathQA & \textbf{Average\textcolor{mygreen}{$\uparrow$}}        \\ \midrule
\multicolumn{1}{c|}{\color[HTML]{9B9B9B}0\%}  & \multicolumn{1}{c|}{\color[HTML]{9B9B9B}Original}  & {\color[HTML]{9B9B9B}5.68}   & {\color[HTML]{9B9B9B}8.35}     & \multicolumn{1}{c|}{\color[HTML]{9B9B9B}7.34}      & {\color[HTML]{9B9B9B} 0.28} & {\color[HTML]{9B9B9B} 0.67} & {\color[HTML]{9B9B9B} 0.67} & {\color[HTML]{9B9B9B} 0.56} & {\color[HTML]{9B9B9B} 0.38} & {\color[HTML]{9B9B9B} 0.78} & {\color[HTML]{9B9B9B} 0.27}  & {\color[HTML]{9B9B9B} 0.52}       \\ \midrule
    \multicolumn{1}{c|}{\multirow{5}{*}{20\%}} & \multicolumn{1}{c|}{SVD}      & 20061           & 20306          & \multicolumn{1}{c|}{18800}          & 0.14  & 0.27  & 0.51  & 0.26  & 0.21  & 0.53  & 0.21  & 0.31 \\
\multicolumn{1}{c|}{}                     & \multicolumn{1}{c|}{FWSVD}     & 1727           & 2152          & \multicolumn{1}{c|}{1511}                 & 0.15  & 0.31  & 0.50  & 0.26  & 0.23  & 0.56  & 0.21  & 0.32 \\
\multicolumn{1}{c|}{}                     & \multicolumn{1}{c|}{ASVD}      & 11.14           & 16.55         & \multicolumn{1}{c|}{15.93}               & 0.25  & 0.53  & 0.64  & 0.41  & 0.27  & 0.68  & 0.24  & 0.43\\ 
\multicolumn{1}{c|}{}                     & \multicolumn{1}{c|}{SVD-LLM}   & 7.94           & 18.05         & \multicolumn{1}{c|}{15.93}                 & 0.22  & 0.58  & 0.63  & 0.43  & 0.29  & 0.69  & 0.24  & 0.44 \\ \cmidrule{2-13} 
\multicolumn{1}{c|}{}                     & \multicolumn{1}{c|}{\sysname}  & \textbf{7.74}  & \textbf{17.35} & \multicolumn{1}{c|}{\textbf{15.03} }   & \textbf{0.28}               & \textbf{0.66}               & \textbf{0.66}               & \textbf{0.46}               & \textbf{0.36}               & \textbf{0.71}               & \textbf{0.25}               & \textbf{0.48}    \\ \midrule
\multicolumn{1}{c|}{\multirow{5}{*}{30\%}} & \multicolumn{1}{c|}{SVD}      & 13103           & 17210          & \multicolumn{1}{c|}{20871}              & 0.13  & 0.26  & 0.51  & 0.26  & 0.21  & 0.54  & 0.22  & 0.30 \\
\multicolumn{1}{c|}{}                     & \multicolumn{1}{c|}{FWSVD}     & 20127           & 11058          & \multicolumn{1}{c|}{7240}               & 0.17  & 0.26  & 0.49  & 0.26  & 0.22  & 0.51  & 0.19  & 0.30  \\
\multicolumn{1}{c|}{}                     & \multicolumn{1}{c|}{ASVD}      & 51           & 70         & \multicolumn{1}{c|}{41}                        & 0.18  & 0.43  & 0.53  & 0.37  & 0.25  & 0.65  & 0.21  & 0.38\\ 
\multicolumn{1}{c|}{}                     & \multicolumn{1}{c|}{SVD-LLM}   & 9.56           & 29.44        & \multicolumn{1}{c|}{25.11}                 & 0.20  & 0.48  & 0.59  & 0.40  & 0.26  & 0.65  & 0.22  & 0.40 \\ \cmidrule{2-13} 
\multicolumn{1}{c|}{}                     & \multicolumn{1}{c|}{\sysname}  & \textbf{9.25}  & \textbf{29.12} & \multicolumn{1}{c|}{\textbf{22.46} }   & \textbf{0.27}               & \textbf{0.63}               & \textbf{0.63}               & \textbf{0.40}               & \textbf{0.30}               & \textbf{0.68}               & \textbf{0.24}               & \textbf{0.45}    \\ \midrule
\multicolumn{1}{c|}{\multirow{5}{*}{40\%}} & \multicolumn{1}{c|}{SVD}      & 52489           & 59977          & \multicolumn{1}{c|}{47774}              & 0.15  & 0.26  & 0.52  & 0.26  & 0.22  & 0.53  & 0.20  & 0.30 \\
\multicolumn{1}{c|}{}                     & \multicolumn{1}{c|}{FWSVD}     & 18156           & 20990          & \multicolumn{1}{c|}{12847}              & 0.16  & 0.26  & 0.51  & 0.26  & 0.22  & 0.53  & 0.21  & 0.30 \\
\multicolumn{1}{c|}{}                     & \multicolumn{1}{c|}{ASVD}      & 1407           & 3292         & \multicolumn{1}{c|}{1109}                  & 0.13  & 0.28  & 0.48  & 0.26  & 0.22  & 0.55  & 0.19  & 0.30\\ 
\multicolumn{1}{c|}{}                     & \multicolumn{1}{c|}{SVD-LLM}   & 13.11           & 63.75       & \multicolumn{1}{c|}{49.83}                 & 0.19  & 0.42  & 0.58  & 0.33  & 0.25  & 0.60  & 0.21  & 0.37 \\ \cmidrule{2-13} 
\multicolumn{1}{c|}{}                     & \multicolumn{1}{c|}{\sysname}  & \textbf{12.39}  & \textbf{55.78} & \multicolumn{1}{c|}{\textbf{41.28} }   & \textbf{0.22}               & \textbf{0.52}               & \textbf{0.61}               & \textbf{0.35}               & \textbf{0.27}               & \textbf{0.62}               & \textbf{0.23}               & \textbf{0.40}    \\ \midrule
\multicolumn{1}{c|}{\multirow{5}{*}{50\%}} & \multicolumn{1}{c|}{SVD}      & 131715           & 87227          & \multicolumn{1}{c|}{79815}             & 0.16  & 0.26  & 0.50  & 0.26  & 0.23  & 0.52  & 0.19  & 0.30 \\
\multicolumn{1}{c|}{}                     & \multicolumn{1}{c|}{FWSVD}     & 24391           & 28321          & \multicolumn{1}{c|}{23104}              & 0.12  & 0.26  & 0.50  & 0.26  & 0.23  & 0.53  & 0.20  & 0.30 \\
\multicolumn{1}{c|}{}                     & \multicolumn{1}{c|}{ASVD}      & 15358           & 47690         & \multicolumn{1}{c|}{27925}               & 0.12  & 0.26  & 0.51  & 0.26  & 0.22  & 0.52  & 0.19  & 0.30\\ 
\multicolumn{1}{c|}{}                     & \multicolumn{1}{c|}{SVD-LLM}   & 23.97          & 150.58        & \multicolumn{1}{c|}{118.57}               & 0.16  & 0.33  & 0.54  & 0.29  & 0.23  & 0.56  & 0.21  & 0.33 \\ \cmidrule{2-13} 
\multicolumn{1}{c|}{}                     & \multicolumn{1}{c|}{\sysname}  & \textbf{19.99}  & \textbf{126.35} & \multicolumn{1}{c|}{\textbf{88.44} }   & \textbf{0.18}               & \textbf{0.42}               & \textbf{0.57}               & \textbf{0.31}               & \textbf{0.23}               & \textbf{0.58}               & \textbf{0.22}               & \textbf{0.36}    \\ \midrule
\end{tabular}
}
\label{tab:dataset_acc}
\end{table*}
}}

\newcommand{\differentllmaccuracyTable}{{
\centering
\captionof{table}{PPL (\textcolor{mygreen}{$\downarrow$}) of three different LLMs -- OPT-6.7B, LLaMA 2-7B, and Mistral-7B -- under 20\% compression ratio on WikiText-2.}
\vspace{2mm}
\resizebox{1\textwidth}{!}{%
\begin{tabular}{c|ccc}
\midrule
\textsc{Method} &OPT-6.7B   & LLaMA 2-7B    & Mistral-7B        \\ \midrule
SVD       & 66275          & 18192         & 159627            \\
FWSVD     & 14559      & 2360          & 6357                    \\
ASVD      & 82         & 10.10         & 13.72              \\ 
SVD-LLM   &16.04       &8.5            &10.21               \\\midrule
\sysname  & \textbf{11.79}         & \textbf{7.70}         &  \textbf{7.57}\\ \midrule
\end{tabular}
}
\label{tab:different_llm_acc}
}}

\newcommand{\largellmaccuracyTable}{{
\centering
\captionof{table}{PPL (\textcolor{mygreen}{$\downarrow$})  of LLaMA-7B, 13B, 30B under 20\% compression ratio on WikiText-2. OOM means out of memory error during the model compression. }
\vspace{-2mm}
\resizebox{1\textwidth}{!}{%
\begin{tabular}{c|ccc}
\midrule
\textsc{Method} &LLaMA-7B   & LLaMA-13B     & LLaMA-30B     \\ \midrule
SVD       & 20061           & 946.31        & 54.11         \\
FWSVD     & 1630            & OOM           & OOM           \\
ASVD      & 11.14            & 6.74          & 22.71         \\
SVD-LLM   & 7.94          & 6.61         & 5.63        \\ \midrule
\sysname    & \textbf{7.74}              & \textbf{6.47}          &  \textbf{5.47} \\ \midrule
\end{tabular}
}
\label{tab:large_llm_acc}
}}

\newcommand{\accuracyTabletwo}{{
\begin{table*}[t]
\centering
\captionof{table}{ 
PPL(\textcolor{mygreen}{$\downarrow$}) and Zero-shot(\textcolor{mygreen}{$\uparrow$}) performance of LLaMA2-7B with Basis Sharing under 20\% to 50\% compression ratios on three language modeling datasets and seven common sense reasoning datasets. The $\mS$ of all language modeling tasks is evaluated with WikiText-2. For reasoning tasks, the $\mS$ of the results outside the bracket is evaluated with WikiText-2, while inside is evaluated with Alpaca.
}
\vspace{-2mm}
\resizebox{1\textwidth}{!}{
\begin{tabular}{cccccccccccc}
\midrule
\multicolumn{1}{c|}{\textsc{Ratio}}           & WikiText-2\textcolor{mygreen}{$\downarrow$} & PTB\textcolor{mygreen}{$\downarrow$} & \multicolumn{1}{c|}{C4\textcolor{mygreen}{$\downarrow$}} & Openb. & ARC\_e & WinoG. & HellaS. & ARC\_c & PIQA & MathQA & \textbf{Average\textcolor{mygreen}{$\uparrow$}}        \\ \midrule
\multicolumn{1}{c|}{\color[HTML]{9B9B9B}0\%}  & {\color[HTML]{9B9B9B}5.47}   & {\color[HTML]{9B9B9B}7.29}     & \multicolumn{1}{c|}{\color[HTML]{9B9B9B}7.29}      & {\color[HTML]{9B9B9B} 0.31} & {\color[HTML]{9B9B9B} 0.76} & {\color[HTML]{9B9B9B} 0.69} & {\color[HTML]{9B9B9B} 0.57} & {\color[HTML]{9B9B9B} 0.43} & {\color[HTML]{9B9B9B} 0.78} & {\color[HTML]{9B9B9B} 0.28}  & {\color[HTML]{9B9B9B} 0.55}       \\ \midrule
\multicolumn{1}{c|}{20\%}                     & 7.77   & 60.00  & \multicolumn{1}{c|}{15.30}    & 0.27 (0.28)               & 0.66 (0.70)              & 0.63 (0.63)              & 0.43 (0.46)               & 0.33 (0.35)              & 0.70 (0.74)              & 0.25 (0.25)               & 0.47 (0.49)    \\
\multicolumn{1}{c|}{30\%}                     & 9.69   & 97.40  & \multicolumn{1}{c|}{23.86}    & 0.26 (0.27)               & 0.58 (0.65)              & 0.62 (0.62)              & 0.38 (0.41)               & 0.27 (0.32)              & 0.66 (0.70)              & 0.23 (0.24)               & 0.43 (0.46)    \\
\multicolumn{1}{c|}{40\%}                     & 13.62  & 195.95 & \multicolumn{1}{c|}{43.89}    & 0.19 (0.21)               & 0.48 (0.57)              & 0.58 (0.57)              & 0.33 (0.36)               & 0.22 (0.27)              & 0.61 (0.66)              & 0.23 (0.23)               & 0.38 (0.41)    \\
\multicolumn{1}{c|}{50\%}                     & 21.3   & 509.30 & \multicolumn{1}{c|}{98.92}    & 0.15 (0.17)               & 0.36 (0.47)              & 0.55 (0.53)              & 0.29 (0.31)               & 0.20 (0.25)              & 0.56 (0.60)              & 0.23 (0.22)               & 0.33 (0.36)    \\ \midrule
\end{tabular}
}
\label{tab:llama2_acc}
\end{table*}
}}

\newcommand{\loraTable}{{
\centering
\captionof{table}{Impact of grouping different numbers of layers on LLaMA-7B under compression ratios from 20\% to 50\%.}
\resizebox{0.9\textwidth}{!}{%
\begin{tabular}{c|cccc}
\midrule
\textsc{\# Layers}     & 20\%       & 30\%      & 40\%      & 50\%     \\ \midrule
1                     & 7.94       & 9.56      & 13.11      & 23.97   \\
2                              & 7.74       & 9.25      & \textbf{12.39}      & \textbf{19.99}   \\
3                              & 7.72       & 9.27      & 12.60      & 20.06   \\
4                              & 7.65      & \textbf{9.18}      & 12.58      & 20.86   \\
5                              & \textbf{7.62}      & 9.19      & 12.81      & 24.45   \\
6                              & 7.64      & 9.20      & 14.13      & 25.40   \\
7                              & 7.67      & 9.24      & 14.64      & 27.30   \\
8                              & 7.75      & 9.49      & 14.60      & 27.92   \\
16                             & 7.95      & 10.58     & 19.72      & 49.11   \\ 
32                             & 7.94      & 9.56      & 30.82      &  85.24  \\ \midrule
\end{tabular}
}
\label{tab:layer_acc}
}}

\newcommand{\loraFinetuneTable}{{
\centering
\captionof{table}{Impact of grouping different numbers of layers on LLaMA-7B under compression ratios from 20\% to 50\% after LoRA Fine-Tuning.}
\resizebox{0.9\textwidth}{!}{%
\begin{tabular}{c|cccc}
\midrule
\textsc{\# Layers}     & 20\%      & 30\%      & 40\%      & 50\%     \\ \midrule
1                     & 7.78      & 9.56      & 10.65     & 13.26   \\
2                              & 7.14      & 7.84      & 8.91      & 10.56   \\
3                              & 7.00      & 7.81      & 9.04      & 10.35   \\
4                              & 7.07      & 7.86      & 9.02      & 10.36   \\
5                              & 6.98      & 8.05      & 9.23      & 10.14   \\
6                              & 6.88      & 8.03      & 9.06      & 10.32   \\
7                              & 6.75      & 7.57      & 9.08      & 10.76   \\
8                              & 6.89      & 7.68      & 9.14      & 10.32   \\
16                             & 7.02      & 7.82      & 9.27      & 11.20   \\ 
32                             & 6.97      & 8.25      & 9.37         & 11.64   \\ \midrule
\end{tabular}
}
\label{tab:layer_lora_acc}
}}

\newcommand{\GPTTable}{{
\centering
\captionsetup{font=small}
\captionof{table}{GPT2 20\% compression ratio compared with Dynamic Tying.}
\vspace{-2mm}
\resizebox{0.48\textwidth}{!}{%
\begin{tabular}{c|ccc}
\midrule
\textsc{Method}     & \# Parm.      & Time      & PPL     \\ \midrule
Dynamic Tying       & 264M (GPT2-XL)      & 13.75h      & 49.37       \\
Basis Sharing       & 94M (GPT2)      & 26.47s      & 43.15       \\ \midrule
\end{tabular}
}
\label{tab:gpt_acc}
}}

\newcommand{\FullFinetuneTable}{{
\centering
\captionof{table}{Result of full parameter fine-tuning by grouping different numbers of layers.}
\vspace{-2mm}
\resizebox{0.4\textwidth}{!}{%
\begin{tabular}{c|cccc}
\midrule
\textsc{\# Layers}     & 20\%      & 30\%      & 40\%      & 50\%     \\ \midrule
2                              & 6.57      & 7.41      & 8.29      & 9.71   \\
4                              & 6.64      & 7.39      & 8.41      & 9.91   \\
8                              & 6.63      & 7.46      & 8.54      & 10.23   \\
16                             & 6.66      & 7.66      & 9.04      & 10.48   \\ 
32                             & 6.67      & 7.90      & 9.24      & 10.94   \\ \midrule
\end{tabular}
}
\label{tab:full_fintune_acc}
}}

\usepackage{microtype}
\usepackage[pdftex]{graphicx}
\usepackage{subfigure}
\usepackage{caption}
\usepackage{xspace}
\usepackage{pifont}
\usepackage{makecell}

\usepackage{subcaption}
\usepackage{booktabs} 
\usepackage{wrapfig}

\PassOptionsToPackage{numbers, compress}{natbib}
\usepackage{iclr2025_conference,times}

\usepackage[utf8]{inputenc} 
\usepackage[T1]{fontenc}    
\usepackage{hyperref}       
\usepackage{url}            
\usepackage{amsfonts}       
\usepackage{nicefrac}       
\usepackage{xcolor}         

\usepackage{amsmath}
\usepackage{amssymb}
\usepackage{mathtools}
\usepackage{amsthm}
\usepackage{multirow}
\usepackage{graphicx}
\usepackage{lscape}
\usepackage{algorithm}
\usepackage{algpseudocode}
\usepackage[capitalize,noabbrev]{cleveref}

\theoremstyle{plain}

\theoremstyle{definition}

\theoremstyle{remark}

\newcommand{\sysname}{\texttt{Basis Sharing}\xspace}

\usepackage[textsize=tiny]{todonotes}
\usepackage{iclr2025_conference,times}

\usepackage{amsmath,amsfonts,bm}

\def\eqref#1{equation~\ref{#1}}

\def\1{\bm{1}}

\def\mB{{\bm{B}}}
\def\mC{{\bm{C}}}

\def\mS{{\bm{S}}}

\def\mU{{\bm{U}}}
\def\mV{{\bm{V}}}
\def\mW{{\bm{W}}}
\def\mX{{\bm{X}}}

\def\mSigma{{\bm{\Sigma}}}

\DeclareMathAlphabet{\mathsfit}{\encodingdefault}{\sfdefault}{m}{sl}
\SetMathAlphabet{\mathsfit}{bold}{\encodingdefault}{\sfdefault}{bx}{n}

\newcommand{\R}{\mathbb{R}}

\usepackage{hyperref}
\usepackage{url}
\usepackage{changes}

\title{Basis Sharing: Cross-Layer Parameter Sharing for Large Language Model Compression}

\author{
\hspace{-5pt}
Jingcun Wang\\
Technical University of Darmstadt\\
\texttt{jingcun.wang@tu-darmstadt.de}
\And
Yu-Guang Chen\\
National Central University\\
\texttt{andyygchen@ee.ncu.edu.tw}
\And
Ing-Chao Lin\\
National Cheng Kung University\\
\texttt{iclin@csie.ncku.edu.tw}
\And 
Bing Li\\
University of Siegen\\
\texttt{Bing.Li@uni-siegen.de}\qquad\quad
\And
Grace Li Zhang\\
Technical University of Darmstadt\\
\texttt{grace.zhang@tu-darmstadt.de}
}

\hypersetup{
  colorlinks,
  linkcolor={red!50!black},
  citecolor={blue!50!black},
  urlcolor={blue!80!black}
}

\iclrfinalcopy 
\begin{document}

\maketitle
\definecolor{mygreen}{HTML}{009901}
\definecolor{myred}{HTML}{A52A2A}
\begin{abstract}
Large Language Models (LLMs) have achieved remarkable breakthroughs. However, the huge number of parameters in LLMs require significant amount of memory storage in inference, which prevents their practical deployment in many applications. 
To reduce memory storage of LLMs, singular value decomposition (SVD) provides a promising solution to approximate weight matrices for compressing LLMs. In this paper, we take a step further to explore parameter sharing across different layers with SVD to achieve more effective compression for LLMs. Specifically, weight matrices in different layers are decomposed and represented as a linear combination of a set of shared basis vectors and unique coefficients. The types of weight matrices and the layer selection for basis sharing are examined when compressing LLMs to maintain the performance. Comprehensive experiments demonstrate that 
Basis Sharing outperforms state-of-the-art SVD-based compression approaches and parameter sharing techniques, especially under large compression ratios. Code is available at: \url{https://github.com/TUDa-HWAI/Basis_Sharing}
\end{abstract}
\section{\textsc{Introduction}}
Large Language Models (LLMs) have revolutionized natural language processing by enabling machines to understand human language more accurately. Although these models have remarkable capabilities, they are computation- and memory-intensive, making their deployment on resource-constrained devices challenging. To address this challenge, model compression has become a widely adopted technique to reduce model size and complexity.

Common compression techniques, such as model distillation \citep{minillm, teachingSmallLM, lion, incontextdis,knowledge2024}, 
pruning \citep{sparsegpt, obc, llmpruner, wanda,classpruning,powerpruning}, 
and quantization \citep{quant-awq,quant-atom,quant-quarot,quant-smoothquant,classquantization}, early-exit \citep{pan2024eetuning,earlyexit} etc.  have been extensively studied. 
While such techniques are effective in many scenarios, these methods often 
require hardware modification  and expensive retraining. Compression techniques based on low-rank approximation with, e.g., Singular Value Decomposition (SVD) \citep{ASVD, fwsvd, svd-llm}, provide a promising alternative since they are not restricted by such constraints. In SVD-based weight compression, a weight matrix in a layer is processed individually by decomposing it into three matrices. By removing small singular values in the decomposed diagonal matrix, the original weight matrix can be approximated with fewer number of weight values.  

Despite the benefits of SVD-based weight compression, the potential of grouping layers for weight approximation and compression has not been explored thoroughly. Since weight matrices in different layers of an LLM might share similarity, parameter sharing across layers can be exploited to further compress weight matrices for LLMs. In sharing parameters across layers, \citet{hay2024dynamic} 
trained a small language model by restricting weight matrices in some layers to be the same. On the one hand, this brute-force method leads to significant performance degradation since weight matrices in different layers should vary to maintain their functionalities. On the other hand, it is impractical to train LLMs from scratch due to limited training data or high training costs. 

Contrary to previous work, in this paper, we use pretrained LLMs to enable weight matrices across layers to share a common set of basis vectors but still retain their different functionalities with unique coefficients. Our method, called Basis Sharing, can compress LLMs effectively. 
In summary, our contributions are as follows:
\begin{enumerate}
    \item We propose to represent weight matrices across different layers in a pretrained LLM with a linear combination of a set of shared basis vectors and coefficients unique to specific layers.  This basis sharing can effectively reduce the number of parameters in LLMs while only affecting the performance of LLMs slightly.

    \item We examine cross-layer basis sharing for different types of weight matrices in LLMs according to the incurred compression errors. The types of weight matrices whose sharing across layers does not incur significant compression error are selected for compressing LLMs. 

    \item For the selected types of weight matrices, we also develop a criterion to group layers to share a set of basis vectors but have individual coefficients to 
    preserve the performance of LLMs.  
    
    \item We conduct extensive experiments on a variety of LLMs, including the LLaMA family \citep{llama, llama2}, OPT-6.7B \citep{opt}, Mistral-7B \citep{mistral}, and GPT-2 \citep{gpt2}. Our Basis Sharing can surpasses the state-of-the-art SVD-based methods in both generation tasks and downstream reasoning tasks without any fine-tuning under compression ratios from 20\% to 50\%. Specifically, compared with state-of-the-art SVD-based compression approaches, Basis Sharing can further reduce the perplexity by up to 25\% on generation tasks and improve accuracy by up to 4\% on downstream reasoning tasks under the same compression ratio. 
\end{enumerate}

\section{\textsc{Related Work}}
\paragraph{Large Language Model Compression} 
 LLM compression techniques include model distillation, pruning and quantization, etc. 
\citet{minillm, incontextdis, teachingSmallLM, lion} successfully applied model distillation to LLM by retraining, 
which incurs high computational cost.
\citet{sparsegpt, obc, wanda, llmpruner} pruned weights that are less sensitive to outliers. However, 
the resulting unstructured weight matrices do not provide meaningful compression benefits on real hardware. Structured pruning techniques, such as 2:4 or 4:8 pruning, can achieve effective compression but restrict a fixed 50\% pruning ratio, which limits flexibility in balancing performance and compression ratio.
\citet{quant-atom, quant-quarot, quant-awq, quant-smoothquant} 
allocated higher quantization bits to weights with larger influence on outliers, 
but it does not reduce the number of parameters, limiting its impact on overall compression.

\paragraph{SVD-based Weight Compression} 
SVD-based weight compression has a flexible compression ratio to maintain performance without retraining. 
\citet{svd1} were the first to apply SVD for neural network compression, and \citet{svd2, svd3} extended this approach to shallow transformer models \citep{transformer}. However, in LLM compression, these methods incur significant errors since they do not consider outliers in activations.
FWSVD \citep{fwsvd} addresses this issue by incorporating the impact of outliers through the Fisher information analysis of weight matrices. However, this method requires gradient information during training process, which is computationally prohibitive for LLMs.
ASVD \citep{ASVD} alleviates this problem by selecting key channels in the weight matrix based on their sensitivity to outliers and minimizing compression error in these channels. While it avoids the need for gradients, ASVD still lacks a direct connection between SVD truncation error and the overall model compression error.
SVD-LLM \citep{svd-llm} improves this by introducing a whitening matrix that captures outlier information, effectively reducing compression error.
\textit{However, all of these methods focus only on compressing individual weight matrices within a single layer, missing the opportunity to exploit weight compression across multiple layers. } 

\paragraph{Parameter Sharing} 
Parameter sharing reduces model size by reusing weight matrices across different layers. Inspired by recurrent neural networks, \citet{share-universal} explored this concept within transformers by restricting all layers in the encoder and decoder to share the same weights. Similarly, \citet{share-2-sub} divided transformer parameters into two groups (attention-related and feedforward-related) and compressed the model by sharing weights within each group.
\citet{share-1} applied selective weight sharing, where specific layers shared the same weights rather than all layers. Beyond direct weight sharing, \citet{share-a1, share-a2} introduced the idea of sharing attention scores between layers. By reusing attention scores, some weight matrices for attention computation could be discarded. Dynamic Tying \citep{hay2024dynamic} determines layer-wise weight sharing during training using reinforcement learning, which is still time-consuming for large LLMs. 
\textit{All of these approaches have been tested only on smaller transformer models and typically require training from scratch or full parameter fine-tuning, which makes them impractical for LLMs. }
\section{Methodology}
\begin{wrapfigure}[12]{r}{0.48\textwidth}
\vspace{-12mm}
    \centering
    \includegraphics[width=1\linewidth]{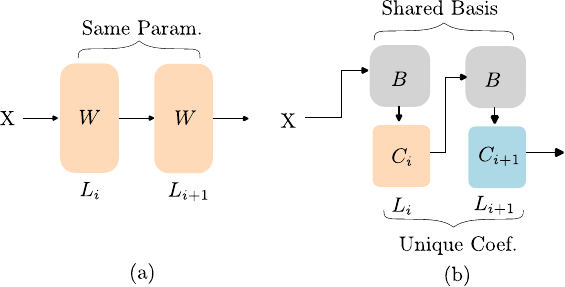}
    \caption{
    (a) Two layers share the same weight matrix in previous work. (b) Two layers share the same basis matrix but have their individual coefficients in our work.
    }
    \label{fig:3.1}
\end{wrapfigure}
Contrary to the previous techniques that require training from scratch and weights in some layers are restricted to be the same during training, 
we adopt a pretrained LLM to explore 
representing weights across different layers with combinations of a set of shared basis vectors and individual coefficients. Since the set of basis vectors can be shared across several layers, the number of parameters in the LLM can thus be reduced effectively. The difference between the previous weight sharing method and our Basis Sharing is illustrated in Figure~\ref{fig:3.1}.

To exploit the cross-layer parameter sharing to compress LLMs, the subsequent subsections address the following challenges: 1) What methodologies can be used to process the weight matrices across layers in an LLM to determine a set of shared basis vectors and individual coefficients? 2) Which types of weight matrices across layers in an LLM can take advantage of parameter sharing without affecting its performance significantly? 3) Which layers can share a set of basis vectors in an LLM without affecting its performance significantly?



\subsection{Representing Weight Matrices across Layers with Combinations of Basis Vectors and Coefficients}\label{section 3.1}

Suppose that we have weight matrices across $n$ layers, denoted as $\displaystyle  \displaystyle \mW^{(1)} \dots \displaystyle \mW^{(n)}, \displaystyle \mW^{(i)} \in \R^{d_1\times d_2}$. To derive a set of shared basis vectors and coefficients to represent such weight matrices, intuitively, such matrices can be horizontally concatenated into one matrix, denoted as $\displaystyle \mW^{}\in \R^{d_1\times nd_2}$, and singular value decomposition (SVD) can be applied to decompose this matrix into three matrices: $\mU, \mSigma, \mV^T$. $\mSigma$ is a $d_1\times nd_2$ diagonal matrix consisting of singular values of $\displaystyle \mW^{}$. 

\begin{figure}[t]
\vspace{-1mm}
    \begin{minipage}{.58\textwidth}
        \centering
        \includegraphics[width=1\linewidth]{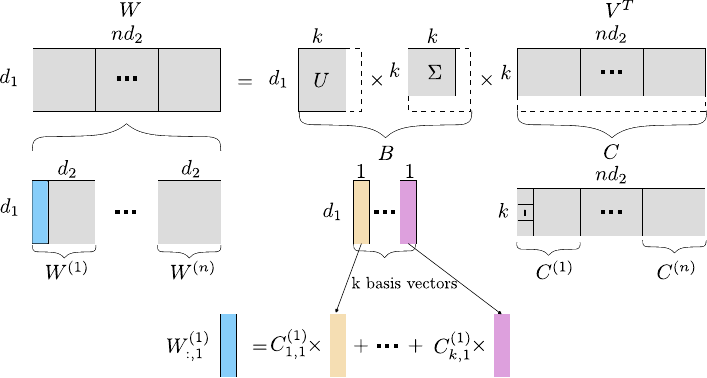}
        \vspace{2mm}
        \caption{Weight matrices across $n$ layers are concatenated horizontally into a weight matrix, which is processed by SVD. The $j^{th}$ column of the original weight matrix in a layer 
        can be represented as a linear combination of $k$ shared basis vectors and coefficients.}
        \label{fig:3.2}
    \end{minipage}%
    \hspace{0.45cm}
    \begin{minipage}{.38\textwidth}
       \centering
        \vspace{2mm}
        \includegraphics[width=0.9\linewidth]{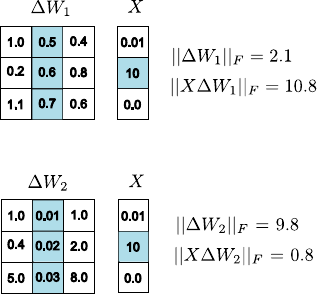}
        \vspace{6mm}
        \caption{
           $\Delta \mW_1$ and $\Delta \mW_2$ are differences with respect to the original weight matrix after compression. $||\Delta \mW_1||_F$ is smaller than $||\Delta \mW_2||_F$, but $||\mX\Delta \mW_1||_F$ is larger than $||\mX\Delta \mW_2||_F$.
        }
        \label{fig:3.3}
    \end{minipage}
\vspace{-3mm}    
\end{figure}

By selecting the top $k$ singular values in $\displaystyle \mSigma$, $\displaystyle \mW^{}$ can be approximated as $\displaystyle \mW^{} \approx \displaystyle \mW_k=\mU_k\mSigma_k\mV^T_k$, where the dimensions of $\displaystyle \mU_k$, $\displaystyle \mSigma_k$ and $\mV^T_k$ are $d_1\times k$, $k\times k$, and $k\times nd_2$, respectively. 
The value of $k$ should be determined to balance the compression ratio and the performance of the compressed LLM (Appendix~\ref{num basis} shows the evaluation of $k$ under a given compression ratio). 
$\displaystyle \mW_k$ can be rewritten as $\displaystyle \mW_k=\mB\mV^T_k$, where $\displaystyle \mB$ is the multiplication result of $\displaystyle \mU_k$ and $\displaystyle \mSigma_k$. We call $\displaystyle \mB$ a basis matrix and a column of $\displaystyle \mB$ is a basis vector, denoted as $\mB_{:,i}$. $\displaystyle \mV^T_k$ can be considered as a coefficient matrix, i.e., $\displaystyle \mV^T_k=\displaystyle \mC$. Accordingly, the $\displaystyle j^{th}$ column of the original  weight matrix 
$\displaystyle\mW^{(i)}$ in the $i^{th}$ layer
can be approximated as a liner combination of 
$k$ basis vectors and individual coefficients as follows. 
\begin{equation}
    \displaystyle \mW^{(i)}_{:, j} \approx  \sum_{m=1}^k \mB_{:,m}\mC^{(i)}_{m, j}.
\end{equation}
where $\mC^{(i)}$ is the coefficient matrix in $i^{th}$ layer.
The process of weight matrix approximation and representation is illustrated in Figure~\ref{fig:3.2}.

In the weight matrix approximation with SVD above, input data, denoted as $\displaystyle \mX$, are not considered. In fact, the result of $\displaystyle \mX\mW$ instead of $\displaystyle \mW$ is used in inference. 
Accordingly, applying SVD directly onto weight matrices 
without incorporating input data might lead to significant computation loss and potentially affect the performance of the LLM. 
Figure~\ref{fig:3.3} illustrates an example, where a weight matrix approximated with SVD leads to a large compression loss in the form of Frobenius loss, denoted as $\displaystyle ||\mX\Delta\mW||_F$. 
Since the second element in the input data affects the computation accuracy significantly, the second column of the weight matrix should be approximated more accurately compared with other columns to reduce the overall computation loss. \citet{ASVD, svd-llm} also pointed out similar results. 


To incorporate the effect of input data into the weight approximation with SVD to maintain the performance of the LLM, we will scale the concatenated weight matrix $\displaystyle \mW$ with a matrix $\displaystyle \mS \in \R^{d_1\times d_1}$ as follows 
\begin{equation}
\displaystyle \mW=\mS^{-1}\mS\mW=\mS^{-1}(\mS\mW).
\end{equation}
The matrix $\displaystyle \mS$ should be evaluated to represent the impact of input data on the weights, so that it can adjust $\displaystyle \mW$ accordingly to reflect the significance of
different input data. To obtain appropriate $\displaystyle \mS$, 
we will adapt the techniques developed in \citet{svd-llm}, where $\displaystyle \mS$ can be evaluated with $\displaystyle \mS(\mS)^{T} = cholesky((\mX)^{T}\mX)$. However, $\displaystyle \mX$ in their technique refers to input data of a layer instead of several layers in our method. To evaluate $\displaystyle \mS$ considering several layers, we will vertically concatenate the input matrices in such layers, denoted as, $\mX^{(1)}, \dots, \mX^{(n)}$,  and compute the $\mS$ with the concatenated $\mX$. 
In our experiments, we use 256 samples from WikiText-2 \citep{wikitext} with each 2048 tokens to evaluate $\mX$, similar to that in \citet{svd-llm}.

Instead of applying SVD directly on the concatenated weight matrix $\displaystyle \mW$, we will decompose $\displaystyle \mS\mW$ with SVD and approximate this scaled weight matrix $\displaystyle \mS\mW \approx \mU^{\prime}_k \mSigma^{\prime}_k \mV^{\prime}_k= \mB^{\prime} \mC^{\prime}$, where $\displaystyle \mB^{\prime}$ and $\displaystyle \mC^{\prime}$ are the revised basis matrix and coefficient matrix, respectively. 
To recover the approximated weight matrix for computation in inference, $\displaystyle \mS^{-1}$ will be multiplied with $\displaystyle \mB^{\prime}$, the result of which will be the final adjusted basis matrix, i.e., 
\begin{equation}
\displaystyle \mW \approx \mS^{-1}\mU^{\prime}_k \mSigma^{\prime}_k \mV^{\prime}_k= \mS^{-1}\mB^{\prime} \mC^{\prime} =\mB^{\prime\prime} \mC^{\prime}, 
\end{equation}
where $\mB^{\prime\prime}$ is the final adjusted basis matrix in our paper.

\subsection{Selection of Weight Matrices in LLMs for Cross-Layer Parameter Sharing}

Modern LLMs are constructed based on the decoder-only transformer architecture. A layer in such an architecture includes several types of weight matrices, which have different functions. $\mW_K$, $\mW_Q$ and $\mW_V$ 
are three types of projection matrices, which are used to generate the key, the query and the value matrices. 
$\mW_O$, another type of weight matrices, further transforms the attention result to build a new representation for an input embedding.
$\mW_{Up}$ and $\mW_{Gate}$(used in  LLaMA and LLaMA2), further types of weight matrices, 
represent this transformation result into a high-dimension embedding. 
Afterwards, $\mW_{Down}$, the last type of weight matrices,  projects the high dimension embedding back to the low dimension embedding. 
The types of weight matrices above have different functions, so that we need to determine which type of weight matrices can take advantage of cross-layer basis sharing with SVD described in Section~\ref{section 3.1} without affecting the performance of the LLM significantly. 

First of all, the type of matrices whose function are to project a high-dimension embedding into a low-dimension embedding such as $\mW_{Down}$ cannot take advantage of the cross-layer parameter sharing. The reason is that after the horizontal concatenation of such matrices, 
the rank of the concatenated matrix will be larger than that of an individual matrix. 
Under the same compression ratio, compressing the concatenated matrix with SVD incurs a larger Frobenius loss than the original weight matrix.  

\begin{figure}[h]
    \centering
    \vspace{-1mm}
    \includegraphics[width=1\linewidth]{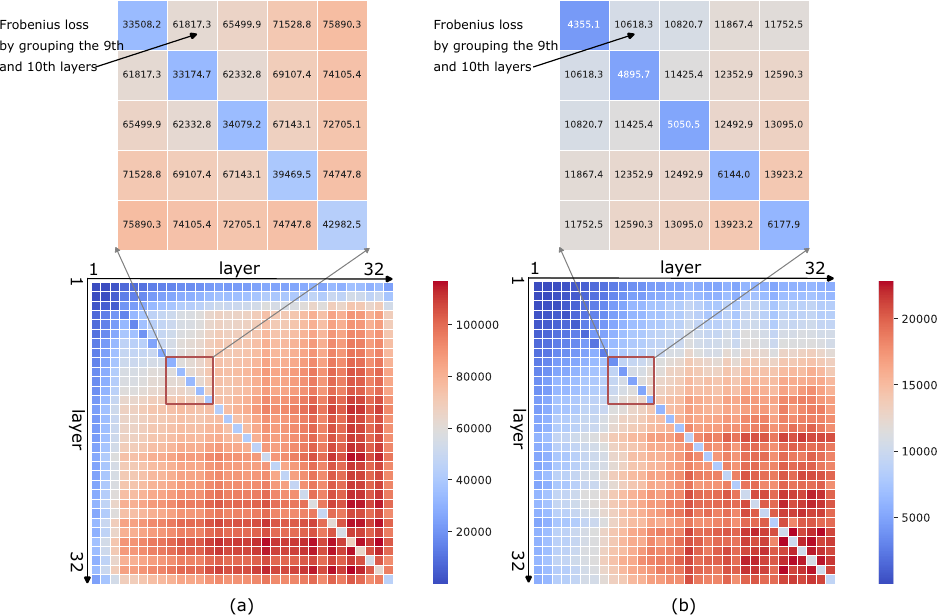}
    \caption{Frobenius loss incurred by basis sharing across any two layers. The number/color in a block represents the resulting Frobenius loss if a basis matrix is shared by
two layers and the numbers in the diagonal direction are obtained by applying SVD to the scaled
weight matrix of a layer directly. (a) Frobenius loss incurred by basis sharing across two layers  for $\mW_K$ in LLaMA2-7B. (b) Frobenius loss incurred by basis sharing across two layers for $\mW_O$ in LLaMA2-7B.}
    \label{fig:3.5}
\end{figure}

For the remaining types of weight matrices including $\mW_K$, $\mW_Q$, $\mW_V$, $\mW_O$, $\mW_{Up}$ and $\mW_{Gate}$, we will determine whether each of them can use cross-layer basis sharing by examining the Frobenius loss resulted from this sharing. 
To explain this concept, 
we use basis sharing  
across two layers for $\mW_K$ in LLaMA2-7B as an example. Assume that we remove small singular values by applying SVD on $\mS^{(i)}_K\mW^{(i)}_K$ 
to achieve a compression ratio of 20\%, where $\mW_K^{(i)}$ is $\mW_K$ matrix in the $i^{th}$ layer ($ i \in [1, 32]$) and  $\mS_K^{(i)}$ is the corresponding $\mS$ matrix for $\mW_K^{(i)}$. The resulting Frobenius loss of each layer under this compression ratio will be evaluated. 
To evaluate the Frobenius
loss incurred by basis sharing, 
we horizontally concatenate $\mW_K^{(i)}$ of the $i^{th}$ layer and $\mW_K^{(j)}$ of the $j^{th}$layer as $\mW_K^{(i, j)}$ where $j\not=i$, $i,j \in [1, 32]$.  SVD is applied on $\mS^{(i, j)}_K\mW_K^{(i, j)}$ to remove small singular values to achieve the same compression ratio, where $\mS^{(i, j)}_K$ is the corresponding S matrix for $\mW_K^{(i, j)}$.
Afterwards, 
we evaluate the incurred Frobenius loss of basis sharing across two layers. 
Similarly, we repeat the process above for $\mW_O$.
The results are illustrated in Figure~\ref{fig:3.5}, where the number/color in a block represents the resulting Frobenius loss if a basis matrix is shared between two layers and the numbers in the diagonal direction are obtained by applying SVD to the scaled weight matrix of a layer directly.

Figure~\ref{fig:3.5}
compares the results of basis sharing for $\mW_K$ and $\mW_O$. 
Basis sharing across two layers for $\mW_K$ can reduce the Frobenius loss. For example,  
when SVD is applied on $\mS_K\mW_K$ for the $9^{th}$ and $10^{th}$ layers separately, the resulting Frobenius loss is evaluated as $33508.2 + 33174.7 = 66682.9$. When the $9^{th}$  and $10^{th}$ layers share a common basis matrix, the Frobenius loss resulting from compression becomes smaller, i.e, $61817.3 < 66682.9$. This indicates that allowing parameter sharing across two layers for $\mW_K$ can enhance computation accuracy. This trend can be seen in $\mW_K$, $\mW_Q$,  $\mW_V$, $\mW_{Up}$ and $\mW_{Gate}$ (Appendix~\ref{heat map} show the results). Accordingly, basis sharing across layers can be applied on such matrices.

On the contrary, basis sharing for $\mW_O$ in $9^{th}$  and $10^{th}$ layers  incurs the increase of the Frobenius loss, i.e., $10618.3 > 4355.1 + 4895.7$. Accordingly, this parameter sharing should not be applied on $\mW_O$ to avoid significant computation loss. For such matrices, we will apply SVD to process the individual matrix in each layer separately. 

\subsection{Selection of Layers for Basis Sharing}
Section \ref{section 3.1} determines which types of weight matrices can be shared across layers. This subsection then determines which layers can share basis vectors to represent such types of weight matrices. To select layers for basis sharing, the basis sharing of such layers should not incur Frobenius loss larger than without sharing.
According to Figure~\ref{fig:3.5}, the group of two adjacent layers leads to smaller Frobenius loss than the sum of the Frobenius loss of two separate layers. Based on this analysis, we will group adjacent layers with the order from the first layer to the last layer. Take a group of two layers as an example. The first layer and the second layer are grouped for basis sharing, followed by the group of the third layer and the fourth layer, etc.

\section{\textsc{Experiments}}
\subsection{\textsc{Settings}}
\paragraph{{Baseline}} 
We compare with the work where SVD-based weight approximation in each individual layer is applied without cross-layer parameter sharing. Such work includes ASVD \citep{ASVD}, FWSVD \citep{fwsvd} and SVD-LLM \citep{svd-llm}. We also compared our method with Dynamic Tying  \citep{hay2024dynamic}, where weights in some layers are restricted to be the same by training from scratch. 
Since this method can only be applied on small language models, only GPT2 \citep{gpt2} was used to compared our method and Dynamic Tying.  

\paragraph{Models and Datasets.} We evaluate our method using several models. For LLMs, many models are evaluated, namely LLaMA family (LLaMA-7B, LLaMA-13B, LLaMA-30B, LLaMA2-7B) \citep{llama, llama2}, OPT-6.7B \citep{opt}, Mistral-7B \citep{mistral}, GPT2. Three language modeling datasets used in our experiment include WikiText-2 \citep{wikitext}, PTB \citep{PTB} and C4 \citep{c4}.  Seven reasoning datasets used in the experiments include OpenbookQA \citep{openbookqa}, WinoGrande \citep{winogrande} HellaSwag \citep{hellaswag}, PIQA \citep{PIQA}, MathQA \citep{mathqa}, ARC-e, ARC-c \citep{arc}. All the reasoning tasks are tested in zero-shot setting with the implementation of LM-Evaluation-Harness framework \citep{eval-harness}.

\paragraph{Implementation details} All of our models are based on the model implemented by the Hugging Face. LLaMA-30B are implemented with FP16, the rest models are implemented with FP32. To evaluate $\mS$, FP64 is used to maintain the computation precision. All experiments are tested on two NVIDIA A100 80GB GPUs. 
$\mS$ is derived through 256 samples from WikiText-2 with 2048 sequence length. 
When the compression 
ratio is 40\% or larger than 40\%
, the incurred compression errors increase, so that the output of a layer as the input of the next layer deviates significantly from its original values. This input deviation affects the evaluations of $\mS$ with $\displaystyle \mS(\mS)^{T} = cholesky((\mX)^{T}\mX)$. To incorporate this input deviation, we update the weights in the next layers for basis sharing with such deviated inputs, similar to that in SVD-LLM.

\subsection{\textsc{Results}}
We evaluate the performance of the proposed cross-layer parameter sharing from four aspects: (a) Performance on generation and reasoning tasks and comparison with state of the art in zero-shot setting. (b) LLM Performance on different LLMs in zero-shot setting. (c) Performance on LLMs with various scales in zero-shot setting. (d) LLM performance with LoRA \citep{lora} fine-tuninng. (e) Comparison with training from scratch for weight sharing across layers. 

\paragraph{Performance on Generation \& Reasoning Tasks}
\accuracyTable

We demonstrate the performance of LLaMA-7B and LLaMA2-7B on ten datasets under different compression ratios from 20\% to 50\%. In evaluating the LLM performance, we group two consecutive layers in the order from the first layer to the last layer to share a basis matrix, while Basis Sharing with more than two layers will be discussed later. Table~\ref{tab:dataset_acc} shows the results of LLaMA-7B. The first three datasets are for text generation tasks and the rest seven datasets are for reasoning tasks. For text generation tasks evaluated by perplexity (PPL), Basis Sharing consistently achieves the lowest PPL among compared with the state-of-the-art methods across all compression ratios and tasks. In reasoning tasks, Basis Sharing 
achieves an average accuracy at least 3\% higher than the state-of-the-art methods. As the compression ratio increases, model performance consistently declines across all the methods due to the incurred larger compression errors. In short, 
Basis Sharing outperforms SVD-LLM due to smaller compression errors as discussed in Section 3. 

\accuracyTabletwo
Table~\ref{tab:llama2_acc} presents the basis sharing results of LLaMA2-7B. For the common reasoning tasks, $\mS$ are evaluated with both WikiText-2 and Alpaca \citep{alpaca} to demonstrate the performance difference. The result outside the bracket is based on the evaluation of $\mS$ with WikiText-2, while the result within the bracket is based on the evaluation of  $\mS$ from Alpaca. It can be seen that LLaMA2-7B is more sensitive to parameter compression, especially on the PTB task. When the compression ratio reaches to 50\%, the PPL of LLaMA2-7B is four times of the PPL of LLaMA-7B, while the performance on the remaining tasks are still comparable.

According to Table~\ref{tab:llama2_acc}, the input dataset from which $\mS$ is derived plays a crucial role in determining performance on common reasoning tasks in zero-shot settings. Generally, the model where $\mS$ is evaluated with Alpaca achieves better accuracy than the model where $\mS$ is evaluated with WikiText-2, especially on ARC\_e under 50\% compression ratio. The accuracy difference can reach 11\%. However, on WinoG. the difference is not obvious, the model where $\mS$ is evaluated with WikiText-2 achieves even higher accuracy under 40\% and 50\% compression ratios.

\paragraph{Performance on Different LLMs}
To evaluate the generalization of Basis Sharing across multiple LLMs, we evaluate its PPL on three distinct models from three LLM families: OPT-6.7B (from the OPT family), LLaMA 2-7B (from the LLaMA family), and Mistral-7B (from the Mistral family). This comparison is conducted under a 20\% compression ratio using the WikiText-2 dataset without any fine-tuning. It can be seen from Table~\ref{tab:different_llm_acc}, Basis Sharing consistently achieves the lowest PPL. Especially for OPT-6.7B and Mistral-7B, Basis Sharing achieves a PPL reduction up to 25\% compared with SVD-LLM.

\begin{figure}[t]
\vspace{-1mm}
    \centering
    \begin{minipage}{.48\textwidth}
      \differentllmaccuracyTable
    \end{minipage}%
    \hspace{0.45cm}
    \begin{minipage}{.48\textwidth}
    \vspace{-1mm}
        \largellmaccuracyTable
    \end{minipage}
\vspace{-3mm}    
\end{figure}

\paragraph{Performance on LLMs with Various Scales}
Basis Sharing can be applied to LLMs with large scales. To demonstrate this, we apply Basis Sharing on LLaMA with three different scales under 20\% compression ratio,  namely LLaMA-7B, LLaMA-13B and LLaMA-30B against the state-of-the-art methods. 
The result is shown in Table~\ref{tab:large_llm_acc}. According to this table, 
Basis Sharing achieves the best performance across all the scales. Since gradient needs to be computed with FWSVD, out of memory error occurs on an A100 GPU. In contrast, Basis Sharing can still be realized with an A100 GPU.

\paragraph{Performance with LoRA Fine-Tuning }
\begin{wrapfigure}[16]{r}{0.48\textwidth}
    \centering
    \vspace{-5mm}
    \includegraphics[width=1\linewidth]{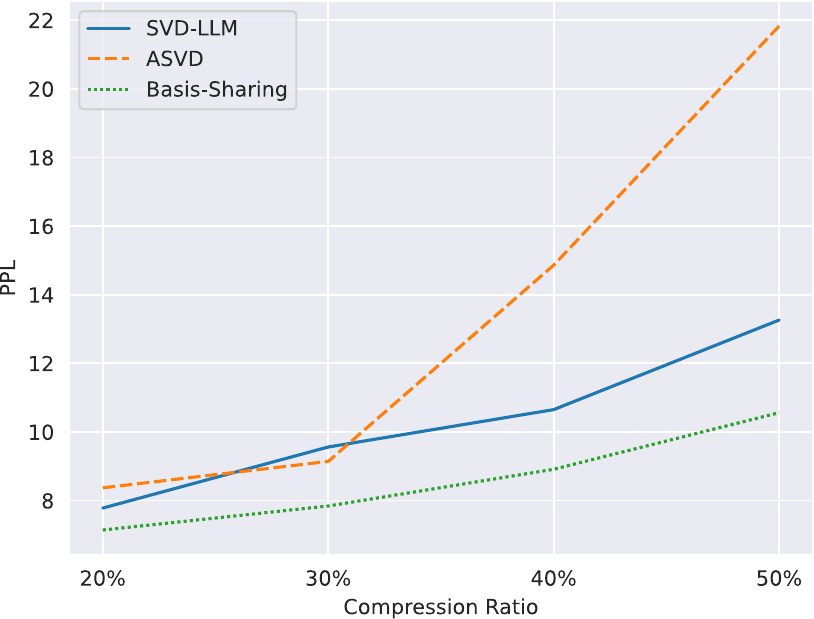}
    \vspace{-6mm}
    \caption{LoRA fine-tuning results of LLaMA-7B under 20\% compression ratio with different compression methods.}
    \label{fig:lora}
\end{wrapfigure}
LoRA \citep{lora} is one of the most promising fine-tuning techniques to recover performance/accuracy. LoRA can also be applied to Basis Sharing to recover performance/accuracy. We used \textit{lora\_r} = 8, \textit{lora\_alpha} = 32, and \textit{learning\_rate} = 1e-4, and used defaults for all other hyperparameters in the Hugging Face PEFT. Each model is fine tuned with WikiText-2 training dataset for two epochs. 

Figure~\ref{fig:lora} shows the result after applying LoRA on LLaMA-7B with WikiText-2. It can be seen from the figure that all compression methods achieve similar PPL under 20\% compression ratio, and PPL difference increases as the compression ratio goes up.  Basis Sharing achieves the lowest PPL when the compression ratio reaches 50\%.

\begin{wraptable}[6]{r}{0.48\textwidth}
\vspace{-4mm}
      \GPTTable
\end{wraptable}

\paragraph{Comparison with Training from Scratch}
Table~\ref{tab:gpt_acc} compares Basis Sharing with Dynamic Tying\citep{hay2024dynamic}, where  parameter sharing is realized by training from scratch. Instead of training from scratch, Basis Sharing leverages pretrained models that have been trained on large datasets and trained with more computational resources. As a result, Basis Sharing achieves fewer parameters, faster compression, and better PPL on WikiText-2 compared to Dynamic Tying.

\subsection{\textsc{Impact of Layer Selection of Basis Sharing on LLM Performance}}
In section 3, we analyzed the change of Frobenius loss when two layers are grouped to share a set of basis vectors. 
In this section, we will demonstrate how grouping more than two consecutive layers affects the LLM performance. 

\begin{figure}[t]
    \centering
    \begin{minipage}{.48\textwidth}
      \loraTable
    \end{minipage}%
    \hspace{0.45cm}
    \begin{minipage}{.48\textwidth}
        \vspace{1mm}
        \loraFinetuneTable
    \end{minipage}
\end{figure}

\paragraph{Impact on LLM Performance in Zero-Shot Setting} 
We grouped different numbers of consecutive layers to examine 
the impact of the number grouped layers on the LLM performance without any fine-tuning. Table~\ref{tab:layer_acc} shows the result. The number in the first column indicates the number of consecutive layers sharing a common basis matrix. For example, 4 means that every four consecutive layers share a basis matrix in the order from the first layer to the last layer. 
Compared with no basis sharing in SVD-LLM (\# LAYERS = 1) under 20\% compression ratio, Basis Sharing achieves a similar performance. 
Grouping four or five layers to share a basis matrix is more reasonable when compression ratio is lower than 30\%, since they have the lowest PPL. Two layers sharing a basis matrix is a good choice when the 
compression ratio is larger than 30\%.

\paragraph{Impact on LLM Performance with LoRA Fine-Tuning}
We also examined the impact of grouping different number of layers on LLM performance after LoRA Fine-Tuning. Table~\ref{tab:layer_lora_acc} 
shows the result. According to this table, 
the performance of LLM can be enhanced compared with that without fine-tuning. 
In addition, this table also shows that after LoRA fine-tuning, grouping layers in LLaMA-7B for Basis Sharing can achieve better performance than that without basis sharing in SVD-LLM (\# LAYERS = 1). 
Even when the number of grouped layer is 32, the performance of Basis Sharing is still better than that without basis sharing in SVD-LLM (\# LAYERS = 1).

\paragraph{Impact on LLM Peformance with Full Parameter Fine-Tuning}

\begin{wraptable}[10]{r}{0.40\textwidth}
\vspace{-4mm}
\FullFinetuneTable
\end{wraptable}

To examine the full potential of the Basis Sharing, we also conducted the full parameter fine-tuning to examine the impact of grouping different numbers of layers on LLM performance. Due to the high computational cost, we only fine tuned the LLaMA-7B on grouping 2, 4, 8, 16, 32 layers, respectively. The differences from LoRA fine-tuning are that we use here \textit{learning\_rate} = 2e-6 and two A100 GPUs. The results of full parameter fine-tuning can be found in Table~\ref{tab:full_fintune_acc}. It can be seen that the performance with full parameter fine-tuning is only a little bit better than the performance with LoRA fine-tuning. The reason could be that WikiText-2 is relatively a small dataset to fine-tune the large model. Directly using this dataset to fine-tune could easily lead to overfitting. Therefore, we reduce the \textit{learning\_rate} from 1e-4 to 2e-6.

\subsection{Performance on Real Hardware}
Basis Sharing not only reduces the memory required for storing parameters, but also enhances inference efficiency on real hardware. 
To demonstrate this advantage, we compared the performance of LLaMA-7B with and without Basis Sharing on a single A100 GPU, using a batch size of 512 and a sequence length of 32 to generate 
one token for each batch.
With this setting, throughput was evaluated as the total number of tokens that can be processed by the model per second. 
\begin{wrapfigure}[17]{r}{0.48\textwidth}
    \vspace{-5mm}
    \centering
        \includegraphics[width=1\linewidth]{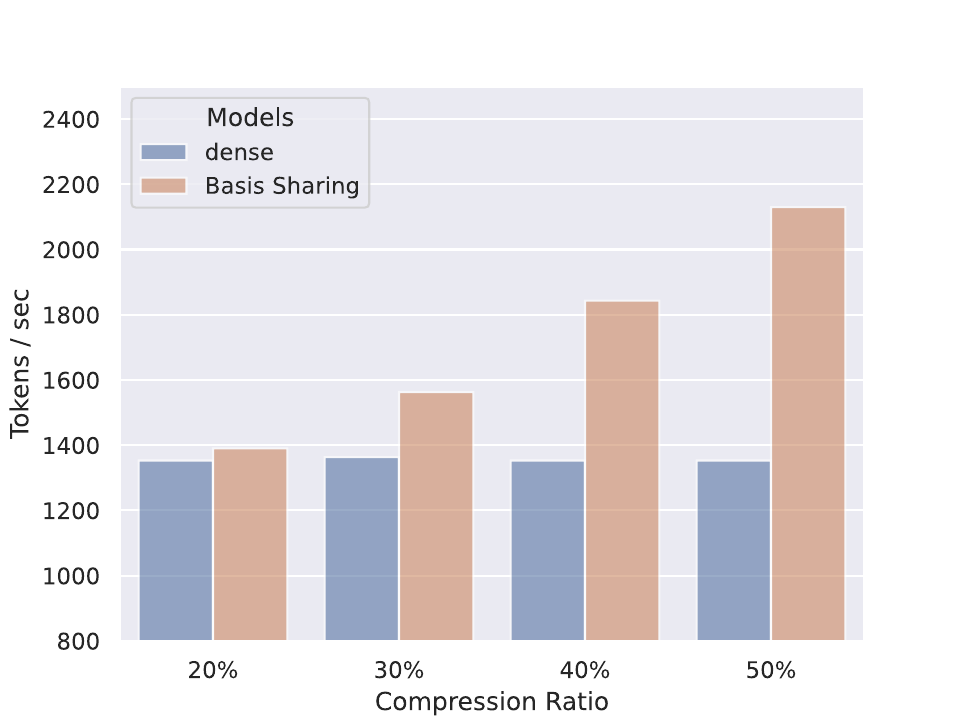}
        \caption{Throughput of dense LLaMA-7B model and the compressed model with Basis Sharing under compression ratios from 20\% to 50\%.}
        \label{fig:mac}
    \vspace{4mm}
\end{wrapfigure}
Figure~\ref{fig:mac} shows the throughput result. It can be seen that as the compression ratio increases, the throughput of model with Basis Sharing also increases. Under 50\% compression ratio, the throughput of Basis Sharing is 1.57 times of the dense model. 

\section{Conclusion}
In this paper, we explore parameter sharing across different layers with SVD to achieve effective compression for LLMs. Specifically, weight matrices in different layers are decomposed and represented as a linear combination of a set of shared basis vectors and unique coefficients. The types of weight matrices and the layer selection for Basis Sharing are examined when compressing LLMs to maintain the performance. Comprehensive experiments demonstrate that  
Basis Sharing outperforms state-of-the-art SVD-based compression approaches, especially under large compression ratios.

\bibliography{iclr2025_conference}

\begin{thebibliography}{48}
\providecommand{\natexlab}[1]{#1}
\providecommand{\url}[1]{\texttt{#1}}
\expandafter\ifx\csname urlstyle\endcsname\relax
  \providecommand{\doi}[1]{doi: #1}\else
  \providecommand{\doi}{doi: \begingroup \urlstyle{rm}\Url}\fi

\bibitem[Amini et~al.(2019)Amini, Gabriel, Lin, Koncel-Kedziorski, Choi, and
  Hajishirzi]{mathqa}
Aida Amini, Saadia Gabriel, Shanchuan Lin, Rik Koncel-Kedziorski, Yejin Choi,
  and Hannaneh Hajishirzi.
\newblock {M}ath{QA}: Towards interpretable math word problem solving with
  operation-based formalisms.
\newblock In \emph{Proceedings of the 2019 Conference of the North {A}merican
  Chapter of the Association for Computational Linguistics: Human Language
  Technologies, Volume 1 (Long and Short Papers)}, pp.\  2357--2367.
  Association for Computational Linguistics, 2019.

\bibitem[Ashkboos et~al.(2024)Ashkboos, Mohtashami, Croci, Li, Jaggi, Alistarh,
  Hoefler, and Hensman]{quant-quarot}
Saleh Ashkboos, Amirkeivan Mohtashami, Maximilian~L Croci, Bo~Li, Martin Jaggi,
  Dan Alistarh, Torsten Hoefler, and James Hensman.
\newblock Quarot: Outlier-free 4-bit inference in rotated llms.
\newblock \emph{arXiv preprint arXiv:2404.00456}, 2024.

\bibitem[Banerjee et~al.(2020)Banerjee, Pal, Mitra, and Baral]{openbookqa}
Pratyay Banerjee, Kuntal~Kumar Pal, Arindam Mitra, and Chitta Baral.
\newblock Careful selection of knowledge to solve open book question answering.
\newblock In \emph{57th Annual Meeting of the Association for Computational
  Linguistics, ACL 2019}, pp.\  6120--6129, 2020.

\bibitem[Bhojanapalli et~al.(2021)Bhojanapalli, Chakrabarti, Veit, Lukasik,
  Jain, Liu, Chang, and Kumar]{share-a2}
Srinadh Bhojanapalli, Ayan Chakrabarti, Andreas Veit, Michal Lukasik, Himanshu
  Jain, Frederick Liu, Yin-Wen Chang, and Sanjiv Kumar.
\newblock Leveraging redundancy in attention with reuse transformers.
\newblock \emph{arXiv preprint arXiv:2110.06821}, 2021.

\bibitem[Bisk et~al.(2020)Bisk, Zellers, et~al.]{PIQA}
Yonatan Bisk, Rowan Zellers, et~al.
\newblock Piqa: Reasoning about physical commonsense in natural language.
\newblock \emph{Proceedings of the AAAI Conference on Artificial Intelligence},
  34:\penalty0 7432--7439, 2020.

\bibitem[Clark et~al.(2018)Clark, Cowhey, Etzioni, Khot, Sabharwal, Schoenick,
  and Tafjord]{arc}
Peter Clark, Isaac Cowhey, Oren Etzioni, Tushar Khot, Ashish Sabharwal, Carissa
  Schoenick, and Oyvind Tafjord.
\newblock Think you have solved question answering? {Try ARC}, the {AI2}
  reasoning challenge.
\newblock \emph{arXiv:1803.05457v1}, 2018.

\bibitem[Dehghani et~al.(2019)Dehghani, Gouws, Vinyals, Uszkoreit, and
  Kaiser]{share-universal}
Mostafa Dehghani, Stephan Gouws, Oriol Vinyals, Jakob Uszkoreit, and Lukasz
  Kaiser.
\newblock Universal transformers.
\newblock In \emph{International Conference on Learning Representations}, 2019.

\bibitem[Frantar \& Alistarh(2022)Frantar and Alistarh]{obc}
Elias Frantar and Dan Alistarh.
\newblock Optimal brain compression: A framework for accurate post-training
  quantization and pruning.
\newblock \emph{Advances in Neural Information Processing Systems},
  35:\penalty0 4475--4488, 2022.

\bibitem[Frantar \& Alistarh(2023)Frantar and Alistarh]{sparsegpt}
Elias Frantar and Dan Alistarh.
\newblock Sparsegpt: Massive language models can be accurately pruned in
  one-shot.
\newblock In \emph{International Conference on Machine Learning}, pp.\
  10323--10337, 2023.

\bibitem[Gao et~al.(2024)Gao, Tow, Abbasi, Biderman, Black, DiPofi, Foster,
  Golding, Hsu, Le~Noac'h, Li, McDonell, Muennighoff, Ociepa, Phang, Reynolds,
  Schoelkopf, Skowron, Sutawika, Tang, Thite, Wang, Wang, and
  Zou]{eval-harness}
Leo Gao, Jonathan Tow, Baber Abbasi, Stella Biderman, Sid Black, Anthony
  DiPofi, Charles Foster, Laurence Golding, Jeffrey Hsu, Alain Le~Noac'h,
  Haonan Li, Kyle McDonell, Niklas Muennighoff, Chris Ociepa, Jason Phang,
  Laria Reynolds, Hailey Schoelkopf, Aviya Skowron, Lintang Sutawika, Eric
  Tang, Anish Thite, Ben Wang, Kevin Wang, and Andy Zou.
\newblock A framework for few-shot language model evaluation, 2024.
\newblock URL \url{https://zenodo.org/records/12608602}.

\bibitem[Golub et~al.(1987)Golub, Hoffman, and Stewart]{svd1}
Gene~H Golub, Alan Hoffman, and Gilbert~W Stewart.
\newblock A generalization of the eckart-young-mirsky matrix approximation
  theorem.
\newblock \emph{Linear Algebra and its applications}, 88:\penalty0 317--327,
  1987.

\bibitem[Gu et~al.(2024)Gu, Dong, Wei, and Huang]{minillm}
Yuxian Gu, Li~Dong, Furu Wei, and Minlie Huang.
\newblock {MiniLLM}: Knowledge distillation of large language models.
\newblock In \emph{The Twelfth International Conference on Learning
  Representations}, 2024.

\bibitem[Hay \& Wolf(2024)Hay and Wolf]{hay2024dynamic}
Tamir~David Hay and Lior Wolf.
\newblock Dynamic layer tying for parameter-efficient transformers.
\newblock In \emph{The Twelfth International Conference on Learning
  Representations}, 2024.

\bibitem[Hsu et~al.(2022)Hsu, Hua, Chang, Lou, Shen, and Jin]{fwsvd}
Yen-Chang Hsu, Ting Hua, Sungen Chang, Qian Lou, Yilin Shen, and Hongxia Jin.
\newblock Language model compression with weighted low-rank factorization.
\newblock In \emph{International Conference on Learning Representations}, 2022.

\bibitem[Hu et~al.(2021)Hu, Shen, Wallis, Allen-Zhu, Li, Wang, Wang, and
  Chen]{lora}
Edward~J Hu, Yelong Shen, Phillip Wallis, Zeyuan Allen-Zhu, Yuanzhi Li, Shean
  Wang, Lu~Wang, and Weizhu Chen.
\newblock Lora: Low-rank adaptation of large language models.
\newblock \emph{arXiv preprint arXiv:2106.09685}, 2021.

\bibitem[Huang et~al.(2022)Huang, Chen, Yu, and McKeown]{incontextdis}
Yukun Huang, Yanda Chen, Zhou Yu, and Kathleen McKeown.
\newblock In-context learning distillation: Transferring few-shot learning
  ability of pre-trained language models.
\newblock \emph{arXiv preprint arXiv:2212.10670}, 2022.

\bibitem[Jiang et~al.(2023{\natexlab{a}})Jiang, Sablayrolles, Mensch, Bamford,
  Chaplot, Casas, Bressand, Lengyel, Lample, Saulnier, et~al.]{mistral}
Albert~Q Jiang, Alexandre Sablayrolles, Arthur Mensch, Chris Bamford,
  Devendra~Singh Chaplot, Diego de~las Casas, Florian Bressand, Gianna Lengyel,
  Guillaume Lample, Lucile Saulnier, et~al.
\newblock Mistral {7B}.
\newblock \emph{arXiv preprint arXiv:2310.06825}, 2023{\natexlab{a}}.

\bibitem[Jiang et~al.(2024)Jiang, Wang, Eldebiky, Yin, Zhuo, Lin, and
  Zhang]{classpruning}
Mengnan Jiang, Jingcun Wang, Amro Eldebiky, Xunzhao Yin, Cheng Zhuo, Ing-Chao
  Lin, and Grace~Li Zhang.
\newblock Class-aware pruning for efficient neural networks.
\newblock In \emph{Design, Automation and Test in Europe Conference and
  Exhibition (DATE)}, 2024.

\bibitem[Jiang et~al.(2023{\natexlab{b}})Jiang, Chan, Chen, and Wang]{lion}
Yuxin Jiang, Chunkit Chan, Mingyang Chen, and Wei Wang.
\newblock Lion: Adversarial distillation of proprietary large language models.
\newblock In \emph{The 2023 Conference on Empirical Methods in Natural Language
  Processing}, pp.\  3134--3154, 2023{\natexlab{b}}.

\bibitem[Lin et~al.(2024)Lin, Tang, Tang, Yang, Chen, Wang, Xiao, Dang, Gan,
  and Han]{quant-awq}
Ji~Lin, Jiaming Tang, Haotian Tang, Shang Yang, Wei-Ming Chen, Wei-Chen Wang,
  Guangxuan Xiao, Xingyu Dang, Chuang Gan, and Song Han.
\newblock Awq: Activation-aware weight quantization for on-device llm
  compression and acceleration.
\newblock \emph{Proceedings of Machine Learning and Systems}, 6:\penalty0
  87--100, 2024.

\bibitem[Lv et~al.(2023)Lv, Zhang, Li, Gan, and Sun]{svd2}
Xiuqing Lv, Peng Zhang, Sunzhu Li, Guobing Gan, and Yueheng Sun.
\newblock Lightformer: Light-weight transformer using svd-based weight transfer
  and parameter sharing.
\newblock In \emph{Findings of the Association for Computational Linguistics:
  ACL 2023}, pp.\  10323--10335, 2023.

\bibitem[Ma et~al.(2023)Ma, Fang, and Wang]{llmpruner}
Xinyin Ma, Gongfan Fang, and Xinchao Wang.
\newblock Llm-pruner: On the structural pruning of large language models.
\newblock \emph{Advances in neural information processing systems},
  36:\penalty0 21702--21720, 2023.

\bibitem[Magister et~al.(2023)Magister, Mallinson, Adamek, Malmi, and
  Severyn]{teachingSmallLM}
Lucie~Charlotte Magister, Jonathan Mallinson, Jakub~Dominik Adamek, Eric Malmi,
  and Aliaksei Severyn.
\newblock Teaching small language models to reason.
\newblock In \emph{The 61st Annual Meeting Of The Association For Computational
  Linguistics}, 2023.

\bibitem[Marcus et~al.(1993)Marcus, Santorini, and Marcinkiewicz]{PTB}
Mitchell~P. Marcus, Beatrice Santorini, and Mary~Ann Marcinkiewicz.
\newblock Building a large annotated corpus of {E}nglish: The {P}enn
  {T}reebank.
\newblock \emph{Computational Linguistics}, 19\penalty0 (2):\penalty0 313--330,
  1993.

\bibitem[Merity et~al.(2016)Merity, Xiong, Bradbury, and Socher]{wikitext}
Stephen Merity, Caiming Xiong, James Bradbury, and Richard Socher.
\newblock Pointer sentinel mixture models, 2016.

\bibitem[Pan et~al.(2024)Pan, Chen, Li, Ding, and Zhou]{pan2024eetuning}
Xuchen Pan, Yanxi Chen, Yaliang Li, Bolin Ding, and Jingren Zhou.
\newblock Ee-tuning: An economical yet scalable solution for tuning early-exit
  large language models, 2024.

\bibitem[Petri et~al.(2023)Petri, Zhang, Chen, Schlichtmann, and
  Li]{powerpruning}
Richard Petri, Grace~Li Zhang, Yiran Chen, Ulf Schlichtmann, and Bing Li.
\newblock Powerpruning: Selecting weights and activations for power-efficient
  neural network acceleration.
\newblock In \emph{Design Automation Conference (DAC)}, 2023.

\bibitem[Qiu et~al.(2024)Qiu, Eldebiky, Li~Zhang, Yin, Zhuo, Schlichtmann, and
  Li]{knowledge2024}
Ruidi Qiu, Amro Eldebiky, Grace Li~Zhang, Xunzhao Yin, Cheng Zhuo, Ulf
  Schlichtmann, and Bing Li.
\newblock Oplixnet: Towards area-efficient optical split-complex networks with
  real-to-complex data assignment and knowledge distillation.
\newblock In \emph{Design, Automation and Test in Europe Conference and
  Exhibition (DATE)}, 2024.

\bibitem[Radford et~al.(2019)Radford, Wu, Child, Luan, Amodei, Sutskever,
  et~al.]{gpt2}
Alec Radford, Jeffrey Wu, Rewon Child, David Luan, Dario Amodei, Ilya
  Sutskever, et~al.
\newblock Language models are unsupervised multitask learners.
\newblock \emph{OpenAI blog}, 1\penalty0 (8):\penalty0 9, 2019.

\bibitem[Raffel et~al.(2019)Raffel, Shazeer, Roberts, Lee, Narang, Matena,
  Zhou, Li, and Liu]{c4}
Colin Raffel, Noam Shazeer, Adam Roberts, Katherine Lee, Sharan Narang, Michael
  Matena, Yanqi Zhou, Wei Li, and Peter~J. Liu.
\newblock Exploring the limits of transfer learning with a unified text-to-text
  transformer.
\newblock \emph{arXiv e-prints}, 2019.

\bibitem[Reid et~al.(2021)Reid, Marrese-Taylor, and Matsuo]{share-2-sub}
Machel Reid, Edison Marrese-Taylor, and Yutaka Matsuo.
\newblock Subformer: Exploring weight sharing for parameter efficiency in
  generative transformers.
\newblock \emph{arXiv preprint arXiv:2101.00234}, 2021.

\bibitem[Sakaguchi et~al.(2021)Sakaguchi, Bras, Bhagavatula, and
  Choi]{winogrande}
Keisuke Sakaguchi, Ronan~Le Bras, Chandra Bhagavatula, and Yejin Choi.
\newblock Winogrande: an adversarial winograd schema challenge at scale.
\newblock \emph{Commun. ACM}, 64\penalty0 (9):\penalty0 99–106, 2021.

\bibitem[Sun et~al.(2024)Sun, Liu, Bair, and Kolter]{wanda}
Mingjie Sun, Zhuang Liu, Anna Bair, and J~Zico Kolter.
\newblock A simple and effective pruning approach for large language models.
\newblock In \emph{The Twelfth International Conference on Learning
  Representations}, 2024.

\bibitem[Sun et~al.(2023)Sun, Zhang, Gu, Lil, and
  Schlichtmann]{classquantization}
Wenhao Sun, Grace~Li Zhang, Huaxi Gu, Bing Lil, and Ulf Schlichtmann.
\newblock Class-based quantization for neural networks.
\newblock In \emph{Design, Automation and Test in Europe Conference and
  Exhibition (DATE)}, 2023.

\bibitem[Takase \& Kiyono(2021)Takase and Kiyono]{share-1}
Sho Takase and Shun Kiyono.
\newblock Lessons on parameter sharing across layers in transformers.
\newblock \emph{arXiv preprint arXiv:2104.06022}, 2021.

\bibitem[Taori et~al.(2023)Taori, Gulrajani, Zhang, Dubois, Li, Guestrin,
  Liang, and Hashimoto]{alpaca}
Rohan Taori, Ishaan Gulrajani, Tianyi Zhang, Yann Dubois, Xuechen Li, Carlos
  Guestrin, Percy Liang, and Tatsunori~B. Hashimoto.
\newblock Stanford alpaca: An instruction-following llama model, 2023.

\bibitem[Touvron et~al.(2023{\natexlab{a}})Touvron, Lavril, Izacard, Martinet,
  Lachaux, Lacroix, Rozi{\`e}re, Goyal, Hambro, Azhar, et~al.]{llama}
Hugo Touvron, Thibaut Lavril, Gautier Izacard, Xavier Martinet, Marie-Anne
  Lachaux, Timoth{\'e}e Lacroix, Baptiste Rozi{\`e}re, Naman Goyal, Eric
  Hambro, Faisal Azhar, et~al.
\newblock Llama: Open and efficient foundation language models.
\newblock \emph{arXiv preprint arXiv:2302.13971}, 2023{\natexlab{a}}.

\bibitem[Touvron et~al.(2023{\natexlab{b}})Touvron, Martin, Stone, Albert,
  Almahairi, Babaei, Bashlykov, Batra, Bhargava, Bhosale, et~al.]{llama2}
Hugo Touvron, Louis Martin, Kevin Stone, Peter Albert, Amjad Almahairi, Yasmine
  Babaei, Nikolay Bashlykov, Soumya Batra, Prajjwal Bhargava, Shruti Bhosale,
  et~al.
\newblock Llama 2: Open foundation and fine-tuned chat models.
\newblock \emph{arXiv preprint arXiv:2307.09288}, 2023{\natexlab{b}}.

\bibitem[Vaswani(2017)]{transformer}
A~Vaswani.
\newblock Attention is all you need.
\newblock \emph{Advances in Neural Information Processing Systems}, 2017.

\bibitem[Wang et~al.(2024{\natexlab{a}})Wang, Li, and Zhang]{earlyexit}
Jingcun Wang, Bing Li, and Grace~Li Zhang.
\newblock Early-exit with class exclusion for efficient inference of neural
  networks.
\newblock In \emph{International Conference on AI Circuits and Systems
  (AICAS)}, 2024{\natexlab{a}}.

\bibitem[Wang et~al.(2024{\natexlab{b}})Wang, Zheng, Wan, and Zhang]{svd-llm}
Xin Wang, Yu~Zheng, Zhongwei Wan, and Mi~Zhang.
\newblock {SVD-LLM}: Truncation-aware singular value decomposition for large
  language model compression.
\newblock \emph{arXiv preprint arXiv:2403.07378}, 2024{\natexlab{b}}.

\bibitem[Wu et~al.(2023)Wu, Kan, Zeng, and Li]{svd3}
Yifan Wu, Shichao Kan, Min Zeng, and Min Li.
\newblock Singularformer: Learning to decompose self-attention to linearize the
  complexity of transformer.
\newblock In \emph{International Joint Conference on Artificial Intelligence},
  pp.\  4433--4441, 2023.

\bibitem[Xiao et~al.(2023)Xiao, Lin, Seznec, Wu, Demouth, and
  Han]{quant-smoothquant}
Guangxuan Xiao, Ji~Lin, Mickael Seznec, Hao Wu, Julien Demouth, and Song Han.
\newblock Smoothquant: Accurate and efficient post-training quantization for
  large language models.
\newblock In \emph{International Conference on Machine Learning}, pp.\
  38087--38099, 2023.

\bibitem[Xiao et~al.(2019)Xiao, Li, Zhu, Yu, and Liu]{share-a1}
Tong Xiao, Yinqiao Li, Jingbo Zhu, Zhengtao Yu, and Tongran Liu.
\newblock Sharing attention weights for fast transformer.
\newblock \emph{International Joint Conference on Artificial Intelligence},
  2019.

\bibitem[Yuan et~al.(2023)Yuan, Shang, Song, Wu, Yan, and Sun]{ASVD}
Zhihang Yuan, Yuzhang Shang, Yue Song, Qiang Wu, Yan Yan, and Guangyu Sun.
\newblock {ASVD}: Activation-aware singular value decomposition for compressing
  large language models.
\newblock \emph{arXiv preprint arXiv:2312.05821}, 2023.

\bibitem[Zellers et~al.(2019)Zellers, Holtzman, Bisk, Farhadi, and
  Choi]{hellaswag}
Rowan Zellers, Ari Holtzman, Yonatan Bisk, Ali Farhadi, and Yejin Choi.
\newblock Hellaswag: Can a machine really finish your sentence?
\newblock In \emph{Proceedings of the 57th Annual Meeting of the Association
  for Computational Linguistics}, pp.\  4791--4800, 2019.

\bibitem[Zhang et~al.(2022)Zhang, Roller, Goyal, Artetxe, Chen, Chen, Dewan,
  Diab, Li, Lin, Mihaylov, Ott, Shleifer, Shuster, Simig, Koura, Sridhar, Wang,
  and Zettlemoyer]{opt}
Susan Zhang, Stephen Roller, Naman Goyal, Mikel Artetxe, Moya Chen, Shuohui
  Chen, Christopher Dewan, Mona Diab, Xian Li, Xi~Victoria Lin, Todor Mihaylov,
  Myle Ott, Sam Shleifer, Kurt Shuster, Daniel Simig, Punit~Singh Koura, Anjali
  Sridhar, Tianlu Wang, and Luke Zettlemoyer.
\newblock Opt: Open pre-trained transformer language models, 2022.

\bibitem[Zhao et~al.(2024)Zhao, Lin, Zhu, Ye, Chen, Zheng, Ceze, Krishnamurthy,
  Chen, and Kasikci]{quant-atom}
Yilong Zhao, Chien-Yu Lin, Kan Zhu, Zihao Ye, Lequn Chen, Size Zheng, Luis
  Ceze, Arvind Krishnamurthy, Tianqi Chen, and Baris Kasikci.
\newblock Atom: Low-bit quantization for efficient and accurate llm serving.
\newblock \emph{Proceedings of Machine Learning and Systems}, 6:\penalty0
  196--209, 2024.

\end{thebibliography}
\bibliographystyle{iclr2025_conference}
\clearpage

\appendix
\section{Appendix}
\subsection{Relation between compression ratio and number of basis  vectors}\label{num basis}
Consider compressing $n$ weight matrices to x\% of their original sizes. Assume each matrix have $d_1$ rows and $d_2$ columns. The number of basis vectors ($k$) can be calculated as follows:
\[
    d_1k + kd_2n = d_1d_2n \times x\% \Rightarrow k = \frac{d_1d_2n\times x\%}{(d_1+d_2n)}
\]

\subsection{Share Error Heat Map}\label{heat map}
The Frobenius loss inccured by basis sharing for $\mW_Q$
, $\mW_V$, $\mW_{Up}$ and $\mW_{Gate}$.

\begin{figure}[h!]
    \centering
    \vspace{-1mm}
    \includegraphics[width=1\linewidth]{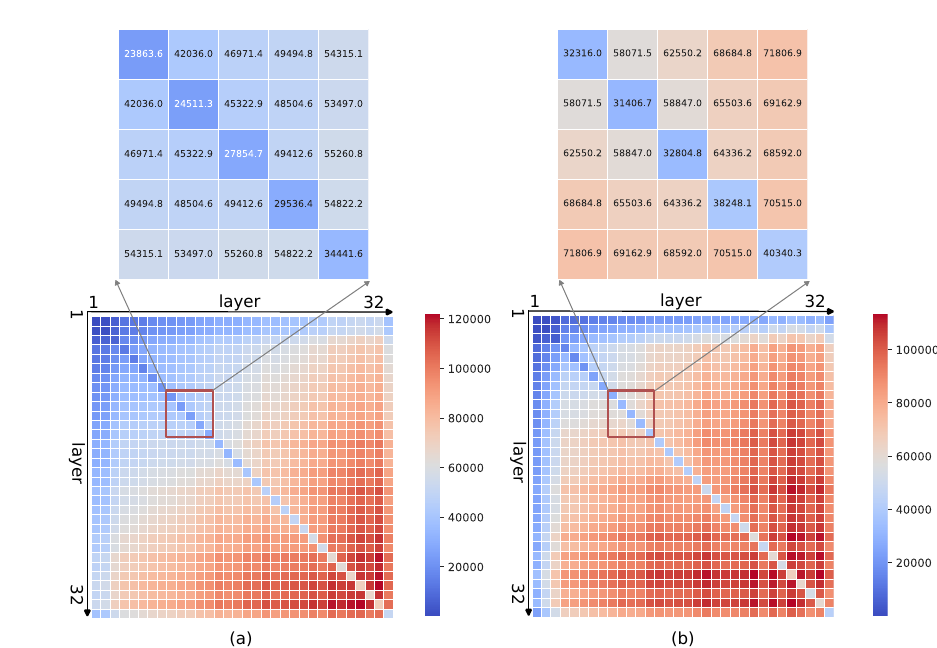}
    \caption{Frobenius loss incurred by basis sharing across any two layers. The number/color in a block represents the resulting Frobenius loss if a basis matrix is shared by
two layers and the numbers in the diagonal direction are obtained by applying SVD to the scaled
weight matrix of a layer directly. (a) Frobenius loss incurred by basis sharing across two layers for $\mW_Q$ in LLaMA2-7B. (b) Frobenius loss incurred by basis sharing across two layers for $\mW_V$ in LLaMA2-7B.}
\end{figure}
\clearpage

\begin{figure}[h!]
    \centering
    \vspace{-1mm}
    \includegraphics[width=1\linewidth]{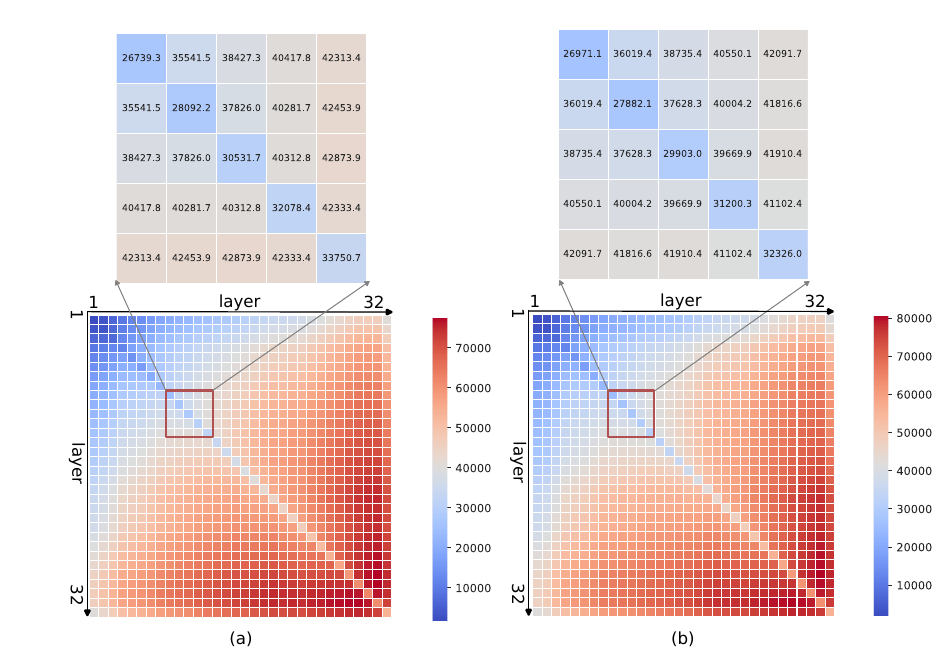}
    \caption{Frobenius loss incurred by basis sharing across any two layers. The number/color in a block represents the resulting Frobenius loss if a basis matrix is shared by
two layers and the numbers in the diagonal direction are obtained by applying SVD to the scaled
weight matrix of a layer directly. (a) Frobenius loss incurred by basis sharing across two layers for $\mW_{Up}$ in LLaMA2-7B. (b) Frobenius loss incurred by basis sharing across two layers for $\mW_{Gate}$ in LLaMA2-7B.}
\end{figure}


\end{document}